\newif\ifdraft\drafttrue
\documentclass{article}
\usepackage{booktabs} 
\usepackage{tikz} 
\usepackage[ruled]{algorithm2e} 

\SetAlFnt{\small}
\SetAlCapFnt{\small}
\SetAlCapNameFnt{\small}
\SetAlCapHSkip{0pt}
\IncMargin{-\parindent}
\usepackage{subcaption}
\usepackage{algorithmic}
\usepackage{enumitem}

\usepackage{blindtext}
\usepackage{hyperref}
\usepackage{url}
\usepackage{comment}
\usepackage{graphicx}
\usepackage{amsmath,amssymb}
\usepackage{amsthm}
\usepackage{fullpage}
\usepackage{bm}
\usepackage{xcolor}
\usepackage[square, comma, numbers,sort&compress]{natbib}
\usepackage{subcaption}
\usepackage{algorithmic}
\usepackage{fullpage}
\usepackage{authblk}
\usepackage{multirow}
\usepackage{multicol}
\usepackage{soul}

\newtheorem{remark}{Remark}

\newcommand{\sj}[1]{\ifdraft{\color{blue}[Shahin: {#1}]}\fi}

\usepackage{sidecap}
\sidecaptionvpos{figure}{c}

\date{}
\title{TorchFL: A Performant Library for \\Bootstrapping Federated Learning Experiments}
\author{Vivek Khimani}
\author{Shahin Jabbari}
\affil{Drexel University\\\{vck29, shahin\}@drexel.edu}

\begin{document}
\maketitle 
\thispagestyle{plain}
\pagestyle{plain}

\begin{abstract}
With the increased legislation around data privacy, federated learning (FL) has emerged as a promising technique that allows the clients (end-user) to collaboratively train deep learning (DL) models without transferring and storing the data in a centralized, third-party server. We introduce \textit{TorchFL} \footnote{The code can be found at \url{https://github.com/vivekkhimani/torchfl} and the documentation can be found at \url{https://torchfl.readthedocs.io/}.}, a performant library for (i) bootstrapping the FL experiments, (ii) executing them using various hardware accelerators, (iii) profiling the performance, and (iv) logging the overall and agent-specific results on the go. Being built on a bottom-up design using PyTorch and Lightning, \textit{TorchFL} provides ready-to-use abstractions for models, datasets, and FL algorithms, while allowing the developers to customize them as and when required. This paper aims to dig deeper into the architecture and design of \textit{TorchFL}, elaborate on how it allows researchers to bootstrap the federated learning experience, and provide experiments and code snippets for the same. With the ready-to-use implementation of state-of-the-art DL models, datasets, and federated learning support, \textit{TorchFL} aims to allow researchers with little to no engineering background to set up FL experiments with minimal coding and infrastructure overhead.
\end{abstract}

\section{Introduction} \label{section_intro}
\label{introduction}
With the rapid advancement of sensing and computing capabilities, the amount of data generated from mobile (client) devices has exponentially increased in recent years \cite{7888916}. The increased volume of data has enabled deep learning  (DL) \cite{lecun2015deep} to become a widely adopted technique to train the computational models on the users' data and actively learn from their browsing patterns. In addition to the availability of the data, the recent research advances in specialized hardware \cite{8662396} have enabled us to achieve astonishing results in ad targeting, language translation, image generation, content recommendations, and a lot more problems that were difficult to solve without a neural network (NN) \cite{dong2021survey}. 

Given the magnitude of data and computational resources required to train these models, the user data is often sent and collected in a centralized server, and the trained model is deployed on the device. While this technique has been effective, the stricter legislation and norms around data privacy make it difficult to use the data for training the DL models \cite{8484769}. As a result, federated learning (FL) has emerged as a promising technique for training models across multiple clients (e.g., edge or mobile devices) without requiring the exchange of locally stored data but only that of the parameters \cite{pmlr-v54-mcmahan17a}. Despite being applied and having displayed initial success with the Google Keyboard \cite{gboard}, FL faces serious hardware and infrastructure challenges before it can provide the same results as the traditional DL techniques \cite{khan2021federated}. 

While part of the research community is dealing with these hardware and infrastructure challenges, most of the community focuses on designing new sampling algorithms, incentive mechanisms, aggregation protocols, defense mechanisms, and all of the other components that are integral to the FL pipeline \cite{9747497}. While working on these problems, it is often a common practice to simulate the FL experiments where the clients, datasets, and FL models are spawned on a single machine and the empirical results are collected. In this paper, we present \textit{TorchFL}, a plug-and-play and performant library for bootstrapping simulated FL experiments.

Running an end-to-end FL experiment using a specific DL framework - PyTorch in this case - consists of various steps including but not limited to the following:
\begin{enumerate}
    \item Selecting and building a DL model compatible with the framework,
    \item Collecting the data and preparing it for training,
    \item Federating the dataset using iid or non-iid configuration,
    \item Writing an aggregation protocol that is used by the server for federation,
    \item Wrapping it all together in a single module,
    \item Setting up the infrastructure to run everything using the accelerated hardware (GPUs, TPUs, etc.).
\end{enumerate}
As the DL models are getting more complex, and constant innovations are being made on the hardware accelerators as well as the tools around the data collection and visualizations, \textit{TorchFL} aims to eliminate such barriers for the FL community by abstracting the hardware, infrastructure, data, and DL implementations to set up the experiments. 
Hence, the main contribution of the \textit{TorchFL} is to provide researchers with no engineering background and novice data scientists a toolkit to set up FL experiments with minimal coding and infrastructure overhead (See Section~\ref{section_related} for a comparison of TorchFL and other FL toolkits). Given that Python and PyTorch are the most popular languages and frameworks used for DL respectively \cite{stanvcin2019overview}, \textit{TorchFL} is a Python library that is built on PyTorch  and can be used by a developer to run an end-to-end experiment by providing the following features:
\begin{enumerate}
    \item Wrappers for state-of-the-art DL models that can be trained in federated or non-federated settings,
    \item Wrappers for the most commonly used state-of-the-art datasets and the ability to automatically create the data shards based on the FL configuration,
    \item Added support for the fine-tuning or feature extraction from the pre-trained DL models, allowing for faster training using federated transfer learning,
    \item Customizable FL layer with the ready-to-use implementation of FL clients, samplers, and aggregators, which can be used to quickly spawn the experiments using the configuration files,
    \item Backward compatibility with the PyTorch Lightning loggers, profilers, hardware accelerators, and the latest DevOps tools to help avoid the implementation and performance overhead for recording and collecting the experimental results.
\end{enumerate}

\begin{figure}[ht!]
  \centering
  \includegraphics[keepaspectratio, width=0.7\columnwidth]{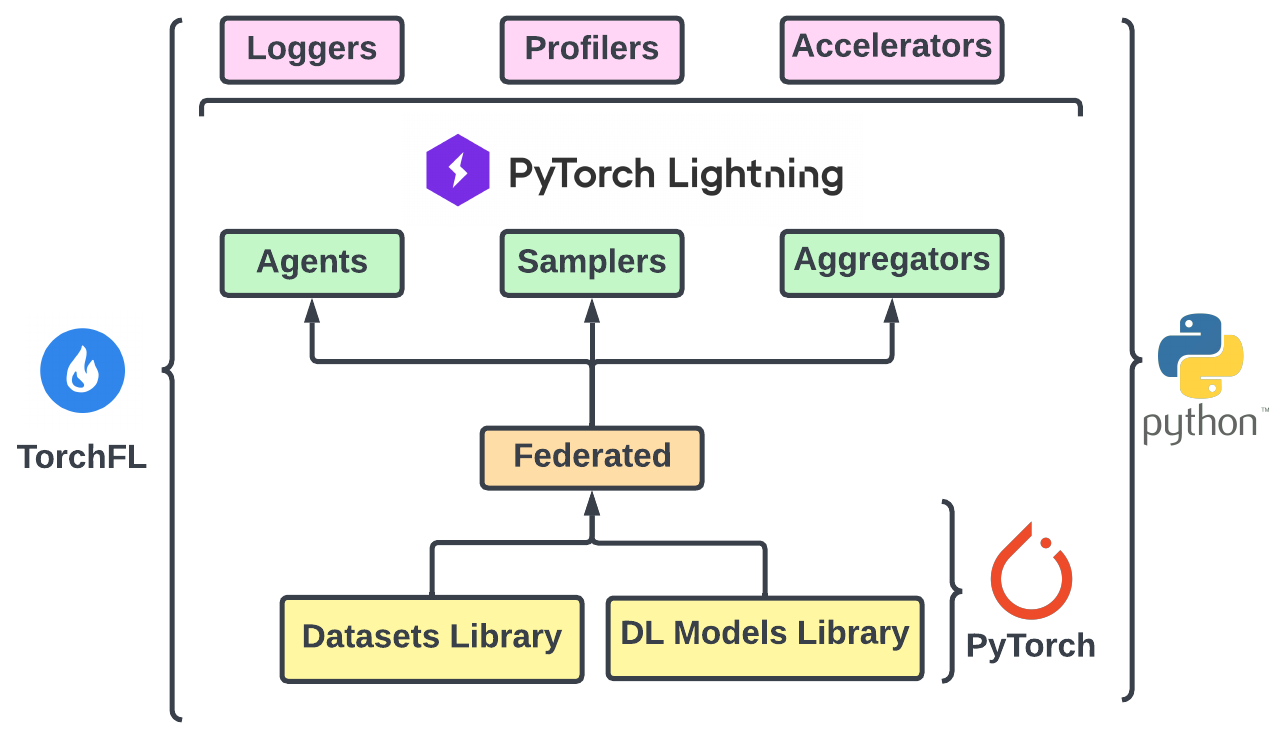}
  \caption{The architectural design of TorchFL.}
  \label{intro_fig}
\end{figure}

In addition to the aforementioned features, abstractions, and ease-of-use, \textit{TorchFL} is implemented using a bottom-up approach as shown in Figure \ref{intro_fig}, which allows the developers to customize every layer and build on top of the library to validate their research and hypothesis.

While \textit{TorchFL} is designed to address the pain points mentioned earlier, it is a relatively new project with its own limitations that we hope to overcome by open-sourcing it and getting feedback from the community. We will dig deeper into these limitations in Section~\ref{section_future}.

The rest of the paper is organized as follows. In Section~\ref{sec:background}, we present background material on FL. Section~\ref{sec:torchfl-arch} discusses the detailed explanation and design decisions for the \textit{TorchFL} architecture. In Section~\ref{section:experiments}, we run FL experiments using  \textit{TorchFL} and provide examples to demonstrate the abstraction, effectiveness, and customizability of the framework. We wrap up by reviewing the existing FL toolkits (Section~\ref{section_related}) and discussing limitations and future work (Section~\ref{section_future}) 

\section{Background on FL}
\label{sec:background}
\begin{figure}[ht!]
  \centering
  \includegraphics[keepaspectratio, width=0.7\columnwidth]{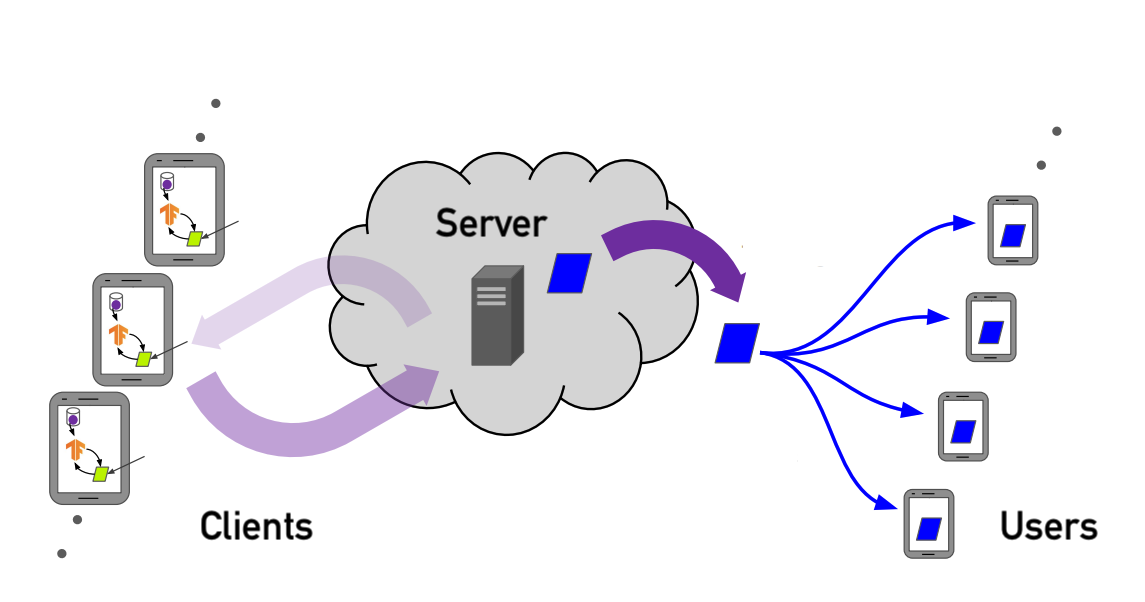}
  \caption{The lifecycle of a \textit{cross-device FL} system with multiple client devices and a single server. The figure is adopted from~\cite{kairouz2021advances}.}
  \label{architecture_fl_bg_fig}
\end{figure}

FL is a machine-learning setting where many clients collaboratively train a model under the orchestration of a central server~\cite{kairouz2021advances}. In this work, we focus on \textit{cross-device FL}, where clients are a very large number of mobile or edge devices with private and locally stored training data. Given a learning objective, the goal of the server is to train a model optimizing for the aforementioned objective by performing updates and aggregation using clients' data without transferring or exchanging any data from the clients. This model can then be deployed on clients as well as new users. The lifecycle of a \textit{cross-device} FL system can be better understood using the visual representation provided in Figure \ref{architecture_fl_bg_fig} \cite{kairouz2021advances}.

Mathematically, the goal in cross-device FL is to optimize a loss function $L$ using a model $M$ parameterized by a vector $W_M \in R^d$ in $T$ rounds via a training dataset $D$, which is distributed and privately stored among a set of $K$ devices or agents $A$ (i.e., $D = D_1 \cup D_2 \cup...\cup D_K$).  At each round $t\in \{1, \ldots, T\}$, a subset of agents, $A^t \subseteq A$, is chosen by the server for training. This process is called sampling. The goal of each sampled agent $i \in A^T$ is to minimize the loss function $L$ over its own privately stored data min-batch of data $D_i^{t}$ at time $t$. The agents minimize the loss $L$ by starting from the global weight vector $W^t_{M}$ at time $t$ and running an algorithm such as stochastic gradient descent (SGD) \cite{mustapha2020overview}. At the end of the update, the agent $i$ obtains a local weight vector $W_{M_i}^{t+1}$ using only their privately stored mini-batch of data samples, computes
\begin{equation} \label{eq:1}
    \delta_i^{t+1} = W_{M_i}^{t+1} - W_M^t
\end{equation}
and sends it to the server. To update the global weight vector $W_M^{t+1}$ for the next round, an \textit{aggregation mechanism} is used over all the collected parameter updates $\Delta^{t+1} = \{\delta_i^{t+1}\}_{i \in A^t}$. A commonly used aggregation mechanism is weighted averaging~\cite{pmlr-v54-mcmahan17a}:
\begin{equation} \label{eq:2}
    W_M^{t+1} =  W_M^t + \sum_{i \in A^t} \Gamma_i^t\delta_i^{t+1} ,
\end{equation}
where $\Gamma_i^t$ is the non-negative weight assigned to agent $i \in A^t$ at time $t$ and $\Sigma_{i \in A^t}\Gamma_i^t=1$. 

In summary, the FL process consists of multiple components including but not limited to \emph{(i)} data distribution among the agents, \emph{(i)} global model selection, \emph{(iii)} agent sampling, \emph{(iv)} local model training and \emph{(v)} aggregation mechanism to update the global model. A practical implementation of a FL system should allow different choices for each of these components. See Section~\ref{sec:torchfl-arch} for more details.

\begin{remark}
The type of data distribution heavily impacts the convergence rate and the performance of FL systems \cite{mustapha2020overview}. There are two common data distributions: (1) \emph{Independent and Identically Distributed (IID)} and (2) \emph{Non-Independent and Identically Distributed (Non-IID)}. Having IID data for the agents means that each batch of data used for a client's local update is statistically identical to a uniformly drawn sample (with replacement) from the entire training dataset $D$ (which is the union of all local datasets for the $K$ agents). In practice, it is unreasonable to assume that every agent has IID data, and hence, throughout the discussion, we heavily focus on the availability of the non-IID datasets.
\end{remark}

\section{Architecture of TorchFL}
\label{sec:torchfl-arch}
Cross-device FL is a multi-step learning process with a lot of implementation, hardware, communication, and systems overhead. Despite the tremendous theoretical success \cite{mustapha2020overview}, research and work are being done to use FL for real-world applications on actual client devices. Yet, as constant research is being done to explore the non-systemic challenges in FL, \textit{TorchFL} aims to provide an end-to-end toolkit for the researchers and developers to quickly bootstrap an FL experiment, orchestrate it on multiple devices, and execute it using the hardware accelerators. As shown in Figure \ref{intro_fig}, \textit{TorchFL} is built using a bottom-up approach to allow the customizability of datasets, models, and the entire FL layer. \textit{TorchFL} is primarily built using Python, PyTorch, and Lightning frameworks and is backward compatible with PyTorch Lightning loggers, profilers, and accelerators. In this section, we explain the motivation behind the design of \textit{TorchFL} and describe each of the layers in \textit{TorchFL}'s architecture.

\subsection{Datasets and Models Libraries} \label{subsection_datasets_models}
As shown in Equation \ref{eq:1}, the local training updates computed by the selected agents are shared with the server to compute the global model parameters as shown in Equation \ref{eq:2}.  In this section, we dig deeper into the design, features, and abstractions offered by the \textit{datasets} and \textit{DL model} offerings in \textit{TorchFL}. As these modules are the first step in setting up an FL experiment, they appear at the lowest layer in Figure \ref{intro_fig}.

    \begin{table}[ht!]
    \caption{Collection of datasets currently supported by \textit{TorchFL} and the availability of IID and non-IID distribution for the agents}
    \label{table:available_datasets}
    \vskip 0.15in
    \begin{center}
    \begin{small}
    \begin{sc}
    \begin{tabular}{lcccr}
    \toprule
    \textbf{Group} & \textbf{Datasets} & \textbf{IID} & \textbf{Non-IID} \\
    \midrule
    \multirow{2}{8em}{CIFAR \cite{krizhevsky2009learning}} & CIFAR-10 & $\surd$ & $\surd$ \\
    & CIFAR-100 & $\surd$ & $\surd$ \\
    \midrule
    \multirow{6}{8em}{EMNIST \cite{cohen2017emnist}}  & By Class & $\surd$ & $\surd$ \\
    & By Merge & $\surd$ & $\surd$\\
    & Balanced & $\surd$ & $\surd$\\
    & Digits & $\surd$ & $\surd$\\
    & Letters & $\surd$ & $\surd$\\
    & EMNIST & $\surd$ & $\surd$\\
    \midrule
    FashionMNIST \cite{xiao2017fashion} & FMNIST & $\surd$ & $\surd$\\
    \bottomrule
    \end{tabular}
    \end{sc}
    \end{small}
    \end{center}
    \vskip -0.1in
    \end{table}

\begin{enumerate}
    \item \textbf{Datamodules}: \textit{TorchFL} currently supports all the datasets listed in Table \ref{table:available_datasets} and provides an end-to-end pipeline for retrieval and distribution of the data. Note that \textit{TorchFL} only currently supports basic computer vision datasets to serve as a proof-of-concept. We hope to add more complex vision-based (Example: ImageNet) and a broader group of datasets in the future.
     All of these data-related features have been wrapped under the \textit{datamodules} module in the codebase and an easy-to-use abstract base class has also been provided for the developers to easily add new datasets. The design of the datamodules interface is explained using a UML diagram in Figure \ref{fig:torchfl_datamodules}.

\begin{figure}[ht!]
  \centering
  \includegraphics[keepaspectratio, width=0.4\textwidth]{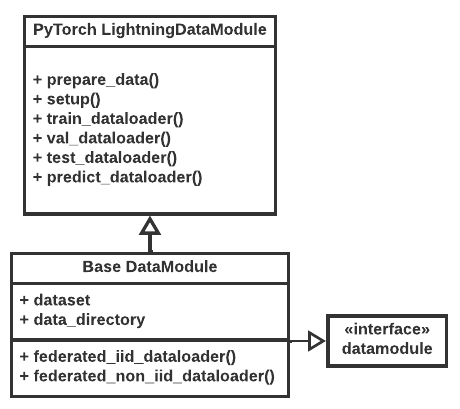}
  \caption{Design and implementation of the datamodules interface in \textit{TorchFL}.}
  \label{fig:torchfl_datamodules}
\end{figure}

    Notably, the abstract base class is inherited and made backward compatible with PyTorch Lightning Datamodule to make full use of their infrastructure and loggers offerings. This design allows us to collect the metadata related to the data distribution alongside all the model-related information which we will see in future sections. As shown in Figure \ref{fig:torchfl_datamodules}, the \textit{datamodules} also provides methods for creating the IID and non-IID distributions for FL. In fact, all of the logic for these methods have been written such that it works out of the box for any new dataset. The non-IID dataloader also provides a parameter named \textit{niid\_factor} which allows the developers to control the intensity of the distribution between the agents. Higher \textit{niid\_factor} would mean a higher imbalance in the labels split between the agents. In summary, any developer willing to add a custom dataset must create a new class, inherit from the \textit{BaseDatamodule}, and only override the relevant methods as required.

    \begin{table}[ht!]
    \caption{Collection of models currently supported by \textit{TorchFL} and the availability of feature extraction and finetuning for every group.}
    \label{table:available_models}
    \vskip 0.15in
    \begin{center}
    \begin{small}
    \begin{sc}
    \begin{tabular}{lcccr}
    \toprule
    \multirow{2}{3em}{\textbf{Models}} & \multirow{2}{5em}{\textbf{Variants}} & \multirow{2}{6em}{\textbf{Feature Extraction}} & \multirow{2}{4em}{\textbf{Fine Tuning}} \\
    & & &\\
    \midrule
    AlexNet    & 1 & $\times$ & $\times$ \\
    DenseNet & 4 & $\surd$ & $\surd$ \\
    LeNet  & 1 & $\times$ & $\times$         \\
    MLP & 1 & $\times$ & $\times$ \\
    MobileNet & 3 & $\surd$ & $\surd$ \\
    ResNet & 9 & $\surd$ & $\surd$ \\
    ShuffleNet & 4 & $\surd$ & $\surd$ \\
    SqueezeNet & 2 & $\surd$ & $\surd$ \\
    VGG & 8 & $\surd$ & $\surd$\\
    \bottomrule
    \end{tabular}
    \end{sc}
    \end{small}
    \end{center}
    \vskip -0.1in
    \end{table}

    \item \textbf{DL Models}: Once the global dataset $D$ has been set up and distributed among the set of $K$ agents $A$, the next step in the FL process is to initialize a global model $M$ which is maintained by a service provider, and assign individual models $M_i$ to every agent $i$. To serve both of these purposes, $TorchFL$ contains a \textit{models} module which contains an extensive library of state-of-the-art DL models as shown in Table \ref{table:available_models}. Every major model has various architectures, a popular example being ResNet \cite{pak2017review} with variants like ResNet18, ResNet34, etc. Although $TorchFL$ currently only supports  the computer vision models, Figure \ref{fig:models_design} explains how the module has been designed with multiple layers of abstraction to prioritize extensibility and ease of adding new models in the future. When trying to add a new model, the developer needs to take the following steps: (i) define the core PyTorch logic at the bottommost layer, (ii) use the templating-based code generator to generate the boilerplate code to ensure compatibility with the datasets, (iii) use the entry points to pass the model and optimizer hyperparameters to generate a trainable model object, and (iv) use the generated models and existing datamodules to initialize a Lightning, compatible trainer. 

\begin{figure}[ht!]
  \centering
  \includegraphics[keepaspectratio, width=0.6\columnwidth]{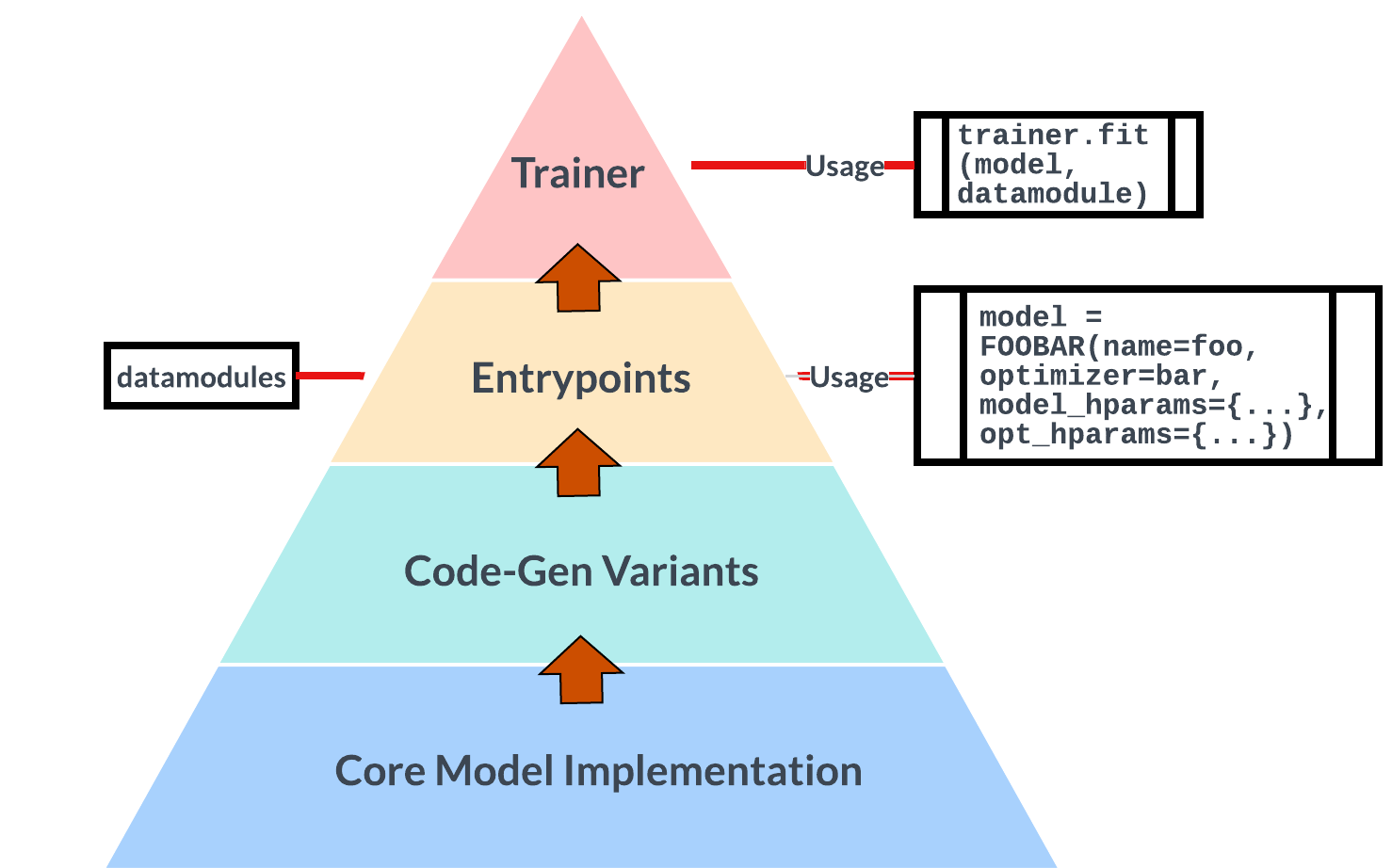}
  \caption{Design and implementation of the model's library in \textit{TorchFL}. The bottom-up approach allows us to implement the core logic and code-generating the boilerplate code for variants. The goal is to have the users utilize the abstraction provided by the trainers at the top.}
  \label{fig:models_design}
\end{figure}

    In addition to the ease of selecting and using state-of-the-art models, \textit{TorchFL} also supports federated transfer learning \cite{ji2021emerging}. Given the millions of trainable parameters, and the amount of time and hardware resources required to achieve convergence, it is unreasonable to expect mobile devices to be able to train these models from scratch \cite{wang2016database}. As such, transfer learning \cite{zhuang2020comprehensive} within FL systems is heavily being studied. The primary idea of transfer learning is to reuse the DL model trained on an existing task for a similar new task. It is analogous to expecting a human proficient at riding bikes to be able to apply or reuse balance and navigation skills while learning to ride motorbikes. \textit{TorchFL} supports the following transfer learning paradigms:
    \begin{enumerate}
        \item \textbf{Finetuning} \cite{ardalan2022transfer}: This approach requires the agents to start with the model parameters pre-trained on an existing task, and retrain them for the new task. Despite the complete retraining, we expect the overall training time to be lower as the initial parameters are not randomly initialized.
        \item \textbf{Feature Extraction}: \cite{zhuang2020comprehensive}: This approach requires the agents to start with the pre-trained model parameters and only retrain the final classification layers instead of the entire model. This technique results in a substantial decrease in the number of trainable parameters and hence accounts for lower training time.
    \end{enumerate}

    Table \ref{table:available_models} contains an archive of all the models available in \textit{TorchFL} that support finetuning and feature extraction. Relevant hyperparameters have been provided in the model entry points to allow the developers to retrieve a finetuned or feature-extracted model on the ImageNet dataset \cite{5206848}. In Section~\ref{section:experiments}, we  demonstrate how transfer learning in FL can help drastically reduce the trainable parameters and training time while achieving optimal performance.
    
\end{enumerate}

\subsection{Federated Module} \label{subsection_federated}
In this section, we discuss how FL modules are designed to work with the datasets and models while being backward compatible with various PyTorch Lightning utilities.

\begin{figure}[t!]
  \centering
  \includegraphics[keepaspectratio, width=0.6 \columnwidth]{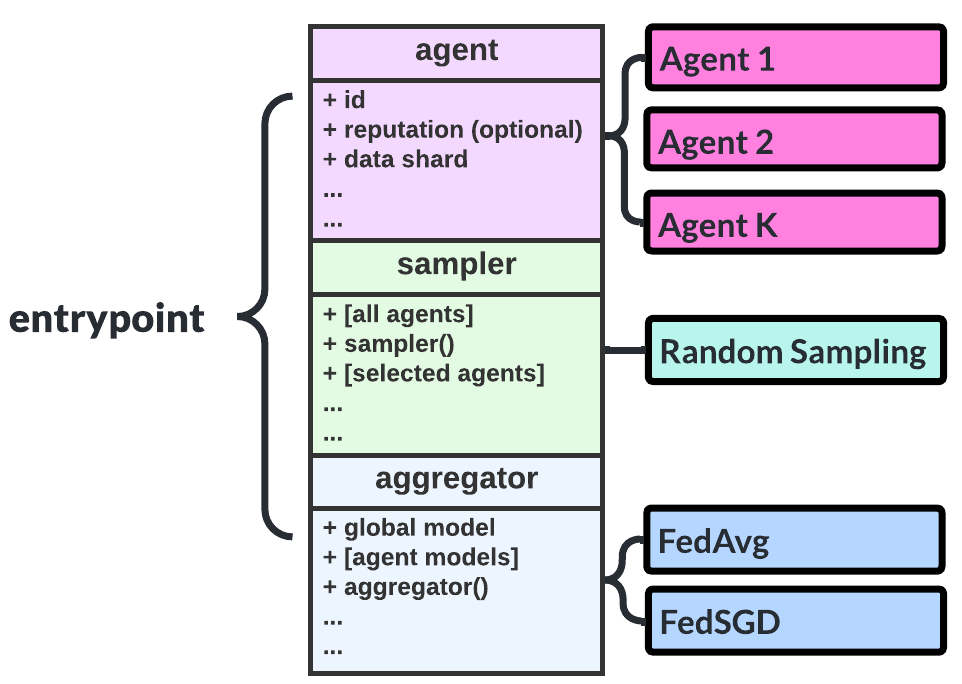}
  \caption{Design and relationships between FL modules in \textit{TorchFL}. As demonstrated in the flowchart, the \textit{agents}, \textit{samplers}, and the \textit{aggregator} are all wrapped by an \textit{entrypoint} object, which is eventually used by the end-user.}
  \label{fig:federated_design}
\end{figure}

\begin{enumerate}
    \item \textbf{Agents}: The primary entity of any FL system is an agent, which is synonymous with a client, mobile, or edge device in the real world. In most of the simulated FL experiments, agents are usually treated as a collection of uniquely identifiable integers which are subject to random sampling and selection for every training round. However, as extensive research is currently being done in areas like reputation-based sampling protocols \cite{yin2021comprehensive}, robust defense mechanisms against model poisoning \cite{9747497}, and game-theoretic incentive systems \cite{zeng2021comprehensive}, we saw value decoupling the \textit{agent} as a separate entity in \textit{TorchFL}. Currently, a \textit{TorchFL} \textit{agent} is identified by a unique identifier, and a shard of the federated dataset when initialized. However, it is designed to be extendable to store more metadata as required. For example, if an incentive mechanism computes the reputation of an agent and uses it to incentivize them for every training round, this can be supported by the \textit{agent} object by adding the relevant fields with relevant datatypes. Lastly, the \textit{agent} object is wrapped under a PyTorch Lightning wrapper which allows us to log and collect all the agent-related metadata, individual agent training performance, and statistics, as and when required.
    \item \textbf{Sampler}: Once the agents are initialized and assigned their shards of the federated data, the \textit{sampler} module is responsible to take the collection of \textit{agent} as an input, and outputting the ones that are selected for training. \textit{TorchFL} currently supports \textit{random sampling} as the baseline algorithm, however, extensive work is being done in exploring the sampling approaches to optimize the learning process \cite{yin2021comprehensive}. In order to add a custom sampling mechanism, the developers need to adhere to the base \textit{sampler} interface and populate the relevant methods with their custom code. Once a new sampler is defined, it can easily be used in any FL experiments via config files to the entry points. 
    \item \textbf{Aggregator}: As explained in Equation \ref{eq:2}, weight or gradient \textit{aggregation} is a core component in the FL pipeline which is responsible for training the global model $M$. Again, \textit{aggregator} has been defined as a separate module in \textit{TorchFL} which takes individual agent models $M_i$ as input and updates the parameters $W_M$ of the global model $M$. Currently, \textit{TorchFL} supports the classic aggregation protocols like \textit{FedAvg} and \textit{FedSGD} \cite{pmlr-v54-mcmahan17a}, but an interface has been provided for the developers to implement their own custom aggregators.
    \item \textbf{Entrypoint}: Once we have the \textit{models}, \textit{dataset}, \textit{agent}, the \textit{sampler}, and an \textit{aggregator}, the last step is to wrap them all to produce a complete FL experiment which can execute on various hardware accelerators, and log the metadata and performance as required. \textit{TorchFL} provides an \textit{entrypoint} module which handles all of these parts and the users are only required to supply their hyperparameters via a config file when initializing the \textit{entrpoint} object. Some good examples of the FL hyperparameters are (i) the number of agents, (ii) the number of global training epochs, (iii) local training epochs, (iv) performance benchmark, (iv) IID or non-IID dataset split, (v) type of sampling mechanism, (vi) type of aggregation protocol, (vii) type of hardware accelerator (TPU, GPU, CPU, or more), (viii) type of logger (CSV, TensorBoard, etc.) and more. As the \textit{entrypoint} module in \textit{TorchFL} automatically abstracts all of these hyperparameters, the developers can only focus on running the required set of experiments.
    See Figure~\ref{fig:federated_design} for more details.
\end{enumerate}

\subsection{Backward Compatibility with PyTorch Lightning} \label{subsection_lightning}
While discussing the design and architecture of various modules in the previous subsections, we have repeatedly emphasized making modules backward compatible with PyTorch Lightning \cite{falcon2019pytorch}. For context, Lightning is a DL framework and wrapper which is built on PyTorch to provide infrastructure support and features without sacrificing the training performance at scale. By making the \textit{TorchFL} modules backward compatible with Lightning, we are able to leverage the following features.

\begin{enumerate}
    \item \textbf{Logging}: It is extremely important to be able to log metadata, metrics, or even configurations while running the experiments. Especially in an FL setting with multiple agents, logging overhead might account for a significant portion of the overall training time, if the loggers are written from scratch in an inefficient manner. As a result, \textit{TorchFL} modules have been written to be backward compatible with Lightning loggers which include but are not limited to CSV, MLFLow, TensorBoard, WeightsAndBiases, and more. Developers can readily configure these loggers without performance or implementation overhead and choose to log anything they need from the experiments.
    \item \textbf{Profiling}: Monitoring the performance or time spent on executing individual components while running massive experiments can be useful. \textit{TorchFL}'s PyTorch Lightning compatibility allows the developers to easily set up the \textit{PyTorchProfiler}, \textit{XLAProfiler} or even build their own profiler as and when required.
    \item \textbf{Hardware Acceleration}: As long as the developers have access to the hardware, this feature allows them to use anything ranging from \textit{CPU}, \textit{GPU}, \textit{HPU}, \textit{IPU}, or even \textit{TPU}. Users can easily choose an accelerator while triggering their experiments or can even switch between accelerators without having to worry about the boilerplate infrastructure setup.
    \item \textbf{Distributed Training}: In a scenario where the developers have access to multiple hardware accelerators, they can leverage the Lightning distributed training \textit{strategies} to parallelize the training process and reduce the overall training time. Notably, FL aggregation falls under an \textit{embarrassingly parallel} algorithm category, and hence distributed training strategies can provide a huge value if chosen meaningfully.
\end{enumerate}

\section{Experiments} \label{section:experiments}
In this section, we present experimental results using \textit{TorchFL} to demonstrate its features. All experiments are performed on the following devices: (i) a virtual machine running on a node with 60GB of RAM, 30 CPU cores, virtualized with the hypervisor running on AMD Epyc cores, and no hardware accelerator, (ii) a virtual machine running on a node with 32GB of RAM, 8 CPU cores, virtualized with the hypervisor running on AMD Epyc cores, and NVIDIA Tesla T4 GPU. For the rest of the section, we will refer to these devices as CPU and GPU, respectively.

\subsection{Usage \& Demo}
Before we dig deeper into the advanced features, we start by demonstrating the basic techniques to use the toolkit and quickly bootstrap the datasets, models, and FL experiments.

\begin{figure}[ht!]
  \centering
  \includegraphics[scale=0.4, width=\columnwidth]{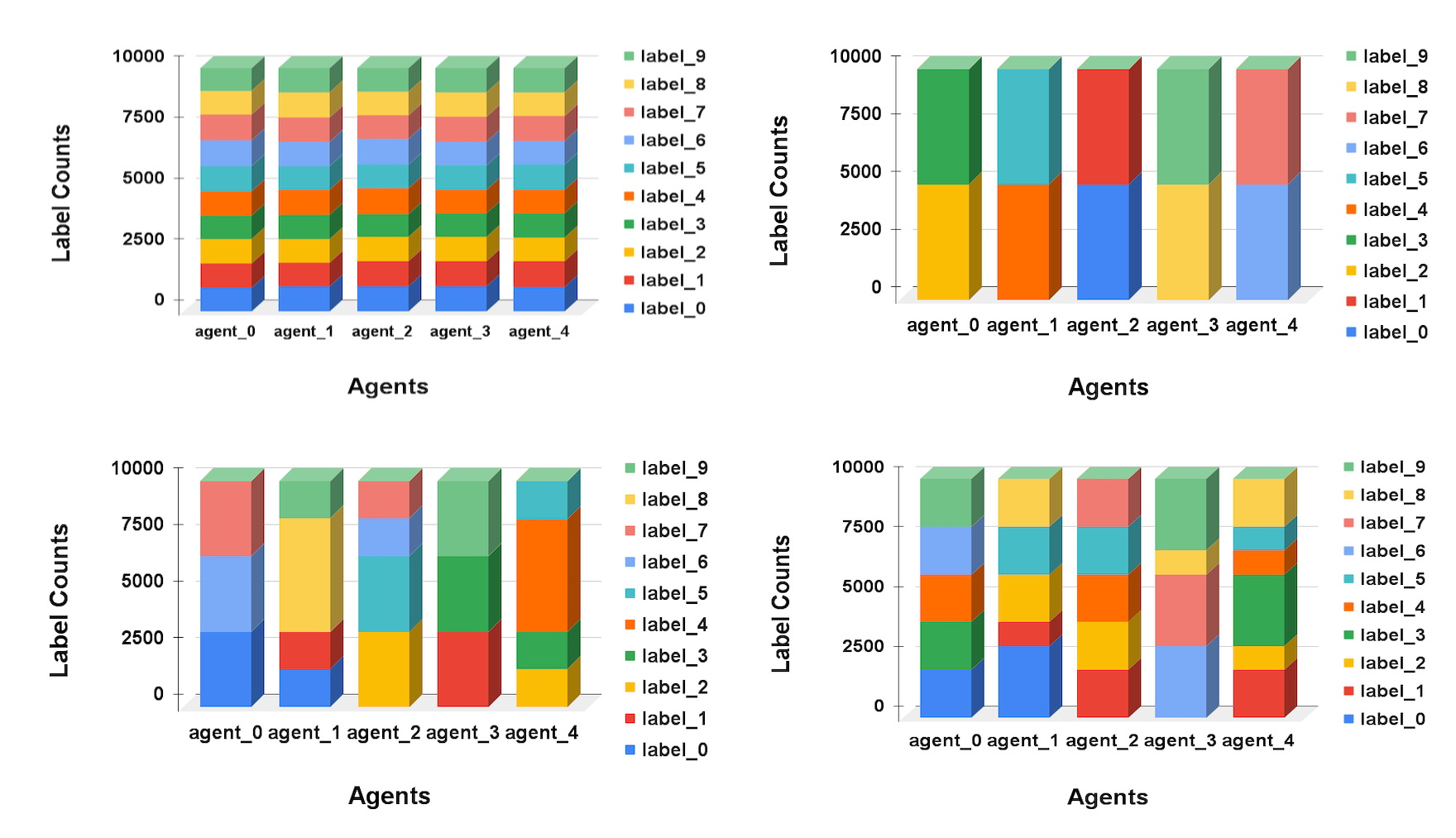}
  \caption{Distribution of labels held by each agent when CIFAR-10 training data (50000) images are split among 5 agents in the following manner: (i) IID, (ii) Non-IID ($niid=1$), (iii) Non-IID ($niid=3$), and (iv) Non-IID ($niid=5$). The number of uniquely held labels by individual agents increases at the niid\_factor increases, with IID being the most evenly balanced configuration.}
  \label{fig:experiments_dataset_iid}
\end{figure}
\subsubsection{Datasets} \label{subsection_experiments_datasets}
We discussed the design of the \textit{datamodules} in Section \ref{subsection_datasets_models} and listed all of our dataset offerings in Table \ref{table:available_datasets}. In this section, we demonstrate the usage of \textit{TorchFL} to promptly set up the CIFAR-10 dataset \cite{Krizhevsky09learningmultiple} and present some visualizations to explain the results. For background, CIFAR-10 is one of the most extensively used datasets in academia given the dimensions and number of total images \cite{obaid2020deep}. It consists of 60000 32$\times$32 color images in 10 classes containing 6000 images each, which are evenly split into 5000 training images and 1000 testing images per class. This means that the dataset consists of 50000 training images and 10000 testing images in total. Each label in the CIFAR-10 dataset can be identified using a unique integer in the range $[0,9]$. For the first experiment, we split the CIFAR-10 training data between $5$ agents, split them using IID and various non-IID configurations provided by \textit{TorchFL}'s datamodule, and visually represent the results in Figure \ref{fig:experiments_dataset_iid}.

All of the datasets-related results shown on the CIFAR-10 dataset with $5$ agents would apply to systems consisting of more agents, or training larger or more complex datasets with more unique labels. Good examples of such datasets are CIFAR-100 \cite{Krizhevsky09learningmultiple} with 100 labels, or even ImageNet \cite{5206848} with 1000 labels. We chose to first demonstrate the results on a smaller experiment configuration to provide more granular visualizations.

\subsubsection{Models}
We discussed the design of \textit{TorchFL}'s \textit{models} in Section \ref{subsection_datasets_models} and listed all of our DL model offerings in Table \ref{table:available_models}. We also discussed the concepts of feature extraction and finetuning in transfer learning to help the models quickly achieve global convergence. In this subsection, we utilize \textit{TorchFl} \textit{model} module to quickly bootstrap various state-of-the-art DL models and provide results to demonstrate the benefits of transfer learning paradigms.

For this set of experiments, we train ResNet152, which is one of the variants of the deep residual networks for image recognition \cite{he2016deep}. In summary, ResNets are a variant of convolutional neural networks (CNN) that democratized the concepts of residual learning and skip connections to enable the training of much deeper models. The model will be trained on the CIFAR-10 dataset, and we experiment on the following settings - (i) training the model from scratch using randomly initialized weights, (ii) finetuning the model which is initialized using the weights pre-trained on the ImageNet dataset, and (iii) feature-extraction, i.e. retraining the classification layers of the model while reusing the pre-trained weights for other layers.

\begin{table}[ht!]
    \caption{Distribution of the trainable, non-trainable, total parameters, and training time in seconds (per epoch) when ResNet152 is trained on CIFAR-10 using an NVIDIA Tesla T4 GPU.}
    \label{table:transfer_learning_parameters}
    \vskip 0.15in
    \begin{center}
    \begin{small}
    \begin{sc}
    \begin{tabular}{lcccr}
    \toprule
    \multirow{3}{4em}{\textbf{Setting}} & \multirow{3}{4em}{\textbf{Train. Param.}} & \multirow{3}{4em}{\textbf{Non-Train. Param.}} & \multirow{3}{4em}{\textbf{Total Param.}} & \multirow{3}{3em}{\textbf{Train. Time}}\\
    & & & &\\
    & & & &\\
    \midrule
    Scratch  & 58.2M & 0 & 58.2M & 1405s\\
    \midrule
    Finetune & 58.2M & 0 & 58.2M & 1380s\\
    \midrule
    \multirow{2}{5em}{Feature. Extract} & 20.5K & 58.1M & 58.2M & 408s\\
    & & & &\\
    \bottomrule
    \end{tabular}
    \end{sc}
    \end{small}
    \end{center}
    \vskip -0.1in
\end{table}

\begin{figure}[ht!]
  \centering
  \includegraphics[keepaspectratio, width=0.7\columnwidth]{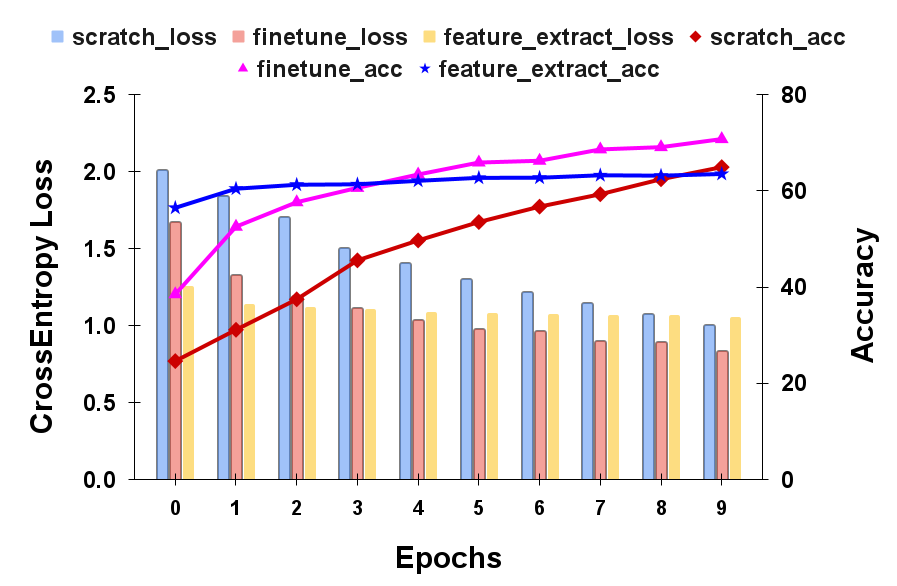}
  \caption{Comparing the validation accuracy and CrossEntropy loss for CIFAR-10 dataset when ResNet152 was trained from scratch, finetuned, and feature-extracted. The comparison between the training time and parameters can be found in Table \ref{table:transfer_learning_parameters}.}
  \label{fig:models_results}
\end{figure}

\begin{figure}[t!]
  \centering
  \includegraphics[scale=0.5, width=\columnwidth]{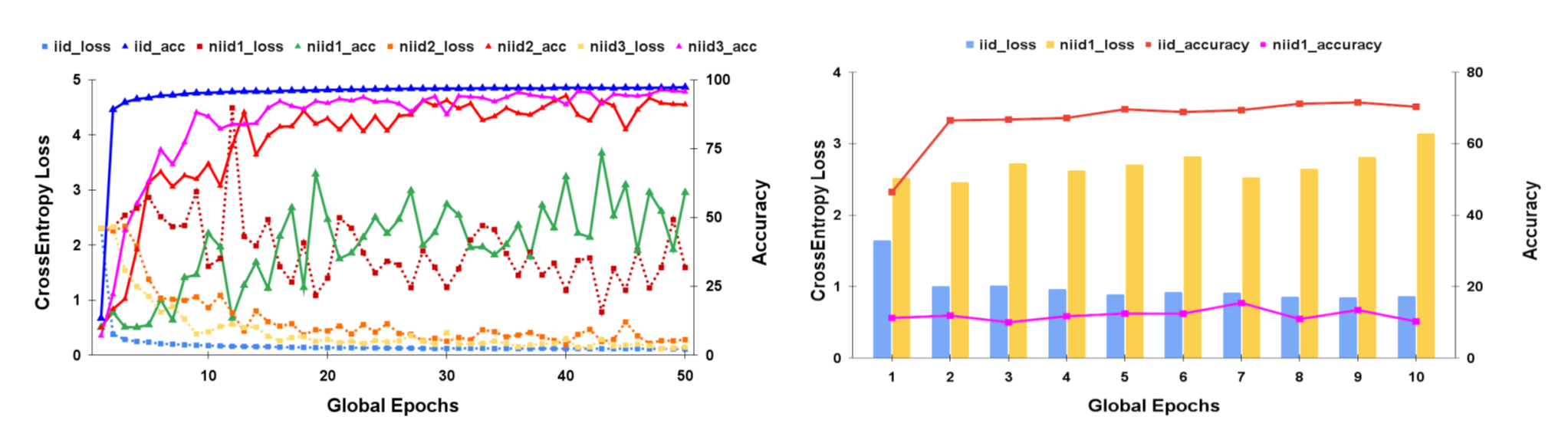}
  \caption{CrossEntropy loss and validation accuracy for the global model when FL experiments are trained with various data distributions and setup using \textit{TorchFL}. (i) 100 agents, 10\% randomly sampled for training, 50 global epochs, 5 local epochs, \textit{FedAvg} aggregation, \textit{LeNet-5} used as a global and local model, (ii) 10 agents, 50\% randomly sampled for training, 10 global epochs, 2 local epochs, aggregated using \textit{FedAvg}, \textit{feature-extracted} \textit{MobileNetV3Small} \cite{howard2019searching} used as a global and local model.} 
  \label{fig:fl_global_model}
\end{figure}

Again, no additional code was written to set up each of these experiments. In fact, we were able to pass the hyperparameters configuration via \textit{TorchFL}'s \textit{model} object and the setup was abstracted under the hood. Before we look at the experimental results, it is worth understanding the primary reason behind the differences in training time and resource utilization between each of these settings. ResNet152 being a massive model, Table \ref{table:transfer_learning_parameters} shows the number of the total, trainable, non-trainable parameters, and training time in seconds (per epoch) when we use either of these training paradigms. We can clearly see how the training time for the feature-extracted model drastically reduces as the number of trainable parameters decrease. As explained in Section \ref{subsection_datasets_models}, the finetuning technique still requires retraining of all the parameters, which means, the training time per epoch is the same as training from scratch. However, as the weights from pre-trained models are used (instead of random initialization), we expect to achieve convergence with a lesser number of epochs, which leads to a decrease in overall training time. The plot in Figure \ref{fig:models_results} shows how the validation accuracy and CrossEntropy loss for ResNet152 vary when trained for 10 epochs using different experimental settings. The plot clearly shows how finetuned and feature-extracted models start with a lower loss because of the pre-trained parameters. Notably, we only trained ResNet152 for 10 epochs as training an entire network on the ImageNet-1K dataset takes roughly more than 3 weeks \cite{he2016deep}. For this work, our goal is to show how \textit{TorchFL} is capable of bootstrapping a model as complicated as ResNet152 and also demonstrate its correctness through the reduction in CrossEntropy \cite{10.5555/3327546.3327555} loss as shown in Figure \ref{fig:models_results}.

\begin{figure}[t!]
  \centering
  \includegraphics[keepaspectratio, width=0.7\columnwidth]{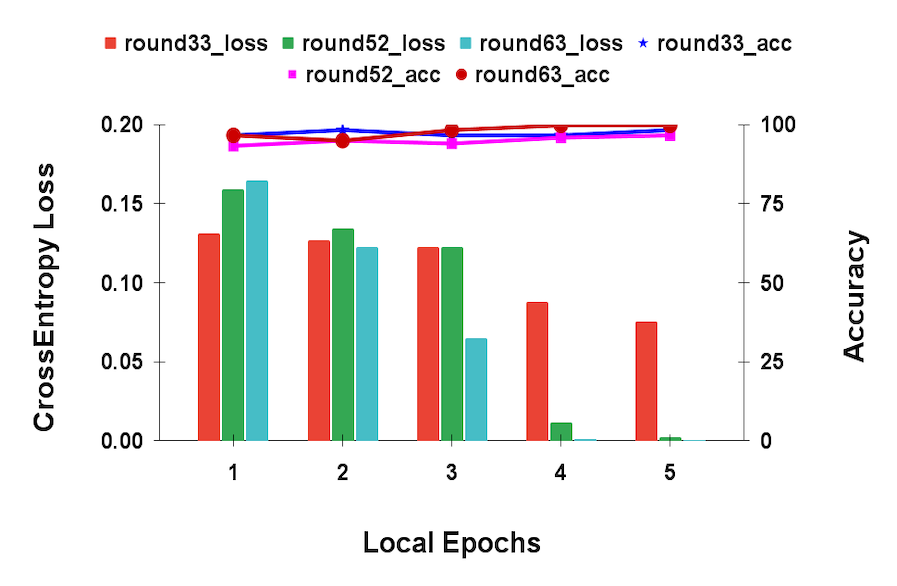}
  \caption{Visualizing the CrossEntropy loss and training accuracy of a randomly selected agent ($id=99$) during the local training when they were selected for training in three different federated global epochs (referred to as rounds in here)}.
  \label{fig:indi_agent_results}
\end{figure}

\subsubsection{Federated Learning}
We discussed the design of the federated module in \textit{TorchFL} in Section \ref{subsection_federated} and saw how the decoupled implementation of \textit{agent}, \textit{sampler}, and \textit{aggregator} are used to provide an \textit{entrypoint} object which is eventually used by the developer. In the previous subsections, we saw how \textit{TorchFL} provides various abstractions around datasets and models. In this section, we will demonstrate an end-to-end FL experiment that utilizes the \textit{datamodule}, \textit{model}, \textit{sampler}, and \textit{aggregator} to train a global model.

For this set of experiments, we will use an MNIST \cite{deng2012mnist}, a database of handwritten digits. It has a training set of 60000 images and a test set of 10000 images. The images are single-channel, B\&W images, with the digits being size-normalized and centered. We choose MNIST over any other dataset for this experiment, as it is relatively easier to achieve convergence \cite{dai2019benchmarking} using a simple CNN like LeNet-5 \cite{726791}. We used \textit{TorchFL}'s datamodules, models, and FL modules to generate an \textit{entrypoint} object, which was used to abstract and generate all the experiments. We primarily aimed to demonstrate FL from scratch and federated transfer learning using different FL configurations. As seen in Figure \ref{fig:fl_global_model}, the trend in the accuracy-loss curves clearly demonstrates learning. Further, we also clearly notice the impact of non-IID data on model convergence. The second plot in Figure \ref{fig:fl_global_model} shows the accuracy-loss trend for the federated transfer learning setup mentioned before. Again, there are multiple combinations of FL parameters that can be used for these experiments and would give us different results. However, our goal is to maximize the number of features used in the demo and hence we limit our experiments such that we can demonstrate FL training from scratch and also via transfer learning respectively.

\subsection{Leveraging PyTorch Lightning Features}
In Section \ref{subsection_lightning} we discussed various benefits provided by the PyTorch Lightning compatibility through various modules in the library. In this section, we present various ways in which we can use the Lightning profilers and loggers to generate meaningful metrics without any implementation or performance overhead.\

\subsubsection{Granular Metrics for Individual Agents}
Collecting granular training and hardware metrics for individual agents, over multiple local and glocal epochs can be tedious and compute inefficient especially while running a massive FL experiment. \textit{TorchFL}'s federated modules compatibility with Lightning loggers automatically logs the training metrics, sampling counts, parameters, device stats, etc. to the configured loggers. Figure \ref{fig:indi_agent_results} shows how \textit{TorchFL} was able to automatically collect the local training metrics of a randomly selected agent, every time they were selected for training.

\subsubsection{Pinpointing the Bottlenecks Using Profilers}
The modules in \textit{TorchFL} being backward compatible with the Lightning profilers, Table \ref{table:experiments_profile_run} presents a sample profiling result that was generated while training LeNet-5 on the MNIST dataset. Note that the results were generated using a \textit{SimpleProfiler} object which only monitors the time for core components. Granular details about the system calls and the bottlenecks can be found using an advanced cProfiler.

    \begin{table}[ht!]
    \caption{Results generated by a \textit{SimpleProfiler} while training LeNet-5 on the MNIST dataset. Only the truncated results have been shown here but more granular results about each of these calls can be achieved via an advanced cProfiler.}
    \label{table:experiments_profile_run}
    \vskip 0.15in
    \begin{center}
    \begin{small}
    \begin{sc}
    \begin{tabular}{lcccr}
    \toprule
    \multirow{2}{4em}{\textbf{Action}} & \multirow{2}{4em}{\textbf{Mean Dur.(s)}} & \multirow{2}{3em}{\textbf{Num Calls}} & \multirow{2}{4em}{\textbf{Total(s)}} & \multirow{2}{4em}{\textbf{Percent.}} \\
    & & & &\\
    \midrule
    \multirow{2}{4em}{Total Run} & \multirow{2}{4em}{-} & \multirow{2}{3em}{$55.7$K} & \multirow{2}{4em}{$36.191$} & \multirow{2}{4em}{$100$} \\
    & & & &\\
    \multirow{2}{4em}{...} & \multirow{2}{4em}{...} & \multirow{2}{3em}{...} & \multirow{2}{4em}{...} & \multirow{2}{4em}{...}\\
    & & & &\\
    \multirow{2}{4em}{LR Sched.} & \multirow{2}{4em}{$0.0006$} & \multirow{2}{3em}{$844$} & \multirow{2}{4em}{$0.1748$} & \multirow{2}{4em}{$0.4748$}\\
    & & & &\\
    \multirow{2}{4em}{Opt. Grad.} & \multirow{2}{4em}{$0.0008$} & \multirow{2}{3em}{$844$} & \multirow{2}{4em}{$0.7654$} & \multirow{2}{4em}{$2.1151$}\\
    & & & &\\
    \bottomrule
    \end{tabular}
    \end{sc}
    \end{small}
    \end{center}
    \vskip -0.1in
    \end{table}

\begin{figure}
  \centering
  \includegraphics[keepaspectratio, width=0.5\textwidth]{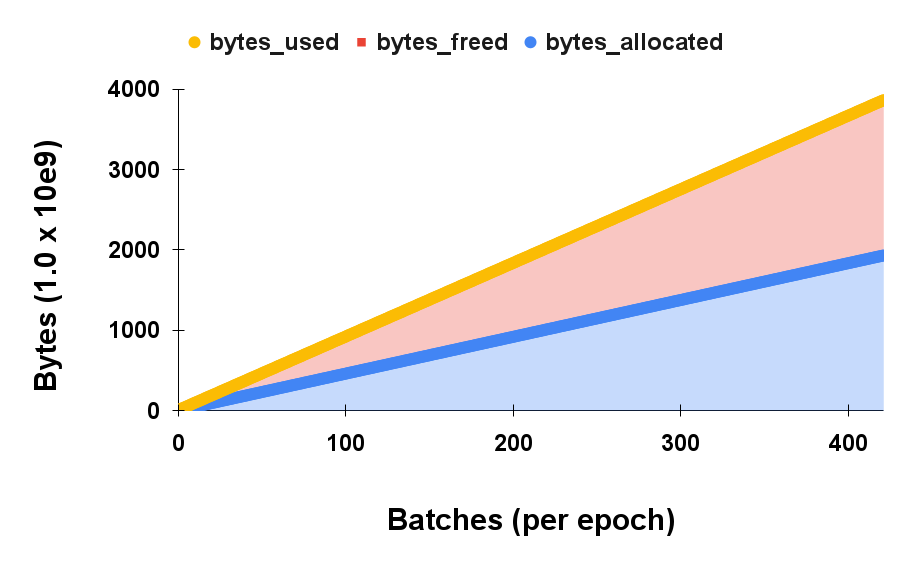}
  \caption{The stacked area chart represents the distribution between the bytes allocated, bytes freed, and bytes used through the batches as we train LeNet-5 for 1 epoch on the MNIST dataset.}
  \label{fig:accelerator_resources}
\end{figure}

\subsubsection{Monitoring the Core Accelerator Resources}
State-of-the-art DL models by themselves require significant memory usage, threads, and pools while training on a hardware accelerator (i.e. GPU). Especially, with FL experiments allocating multiple copies of models for individual agents, this usage is only expected to shoot upwards. As a result, \textit{TorchFL}'s ability to monitor the hardware used for individual agent models provides significant value to the developers. Figure \ref{fig:accelerator_resources} is an example of how \textit{TorchFL} can monitor the memory usage on the hardware accelerator through the training process. Some other useful metrics recorded by \textit{TorchFL} include thread pool allocation and thread usage within the individual pools.

\section{Related Work}
\label{section_related}
In this section, we discuss, compare, and analyze various open-source toolkits in the FL space. The goal of this section is to provide a broad, high-level overview of tools available to the best of our knowledge, and analyze how they compare to \textit{TorchFL}, or can even be used in tandem with \textit{TorchFL}.

Lately, a few open-source offerings have emerged in the FL space with one of the most promising ones being FedML \cite{he2020fedml}. With a multi-layer architecture, and beta offerings for iOS, Android, and IoT devices, FedML aims to provide a complete ecosystem to implement an end-to-end, real-world FL system. However, due to a lack of a standardized underlying framework (eg. PyTorch), setting up FL experiments on FedML still require overhead costs of implementing the models, datasets, and FL utilities that are compatible with the core API offering. On the other hand, with the features like models and dataset offerings, granular metric collection, and customizable interfaces, \textit{TorchFL} aims to solely optimize for the ease of bootstrapping the FL experiments and integrate state-of-the-art research (samplers, aggregators, etc.) with minimum overhead.

LEAF \cite{caldas2018leaf} is another promising framework that aims to open-source and benchmark suitable datasets for FL settings. A few of their notable dataset offerings include Reddit data for language modeling, Shakespeare manuscripts for next-character prediction, and more. We believe these datasets can be readily ported into a \textit{TorchFL} \textit{datamodule} interface and can readily be trained and tested with our DL model offerings.

Moreover, Tensorflow-Federated (TFF) \cite{tff}, PySyft \cite{ryffel2018generic}, FATE \cite{JMLR:v22:20-815}, FLUTE \cite{msflute}, FedScale \cite{fedscale}, and FLOWER \cite{beutel2020flower} are among the few other notable toolkits in the FL space. TFF has extensive support for various aggregation protocols, analytics, profiling, backends, and more, but it's built on top of Tensorflow and lacks standardized support for data modules as of yet. While  PySyft is an actively-maintained and robust offering, their major contributions are ready-to-use algorithms for privacy-preserving techniques like differential privacy \cite{hassan2019differential} and encrypted computation \cite{saleh2016processing}. PySyft's FL offering is the infrastructure to being able to set up privacy-aware agents and service providers as separate entities and able to develop custom PyTorch models, and aggregation protocols on top of it while ensuring private exchange of the computational structures, i.e. tensors. Next, FATE is a group of multiple FL frameworks that aim to let developers deploy their FL workflows on industrial-grade infrastructure and allow multiple organizations to potentially engage in the FL training process. In addition to their primary FL offering, they offer multiple tools for serving FL models on various environments and utilities for exchanging the data over the network. Lastly, FLUTE~\cite{msflute} is another promising toolkit that's built for PyTorch and aims to streamline the FL experiments and simulations. It's meant to be configurable using YAML file format and still requires the developers to build their own PyTorch models.

\section{Limitations and Future Work}
\label{section_future}
This section contains a high-level overview of the plans of maintaining, improving, and refactoring \textit{TorchFL} in the future. As mentioned before, \textit{TorchFL} is a relatively new project with its own limitations and we plan to overcome them by open-sourcing the project and getting feedback from the community. As there are multiple directions in which this work can be extended, we restrict the scope of this discussion and attempt to summarize it in the following manner.

\subsection{Exploring Performance Enhancements}
As mentioned earlier in Section \ref{subsection_lightning}, Lightning distributed training strategies can be used in \textit{TorchFL} to reduce the training time if developers have access to multiple hardware accelerators. In addition, the recent release of PyTorch C++ API and the related tools (TorchScript, ATen, etc.) now allows developers to leverage Python and PyTorch to write custom kernels using CUDA\cite{8748495}, that is backward compatible with the original PyTorch implementations. Once we ensure the stability of the current features and offerings, we are inclined to explore the possibilities of high-performance frontiers on \textit{TorchFL}.

\subsection{Added Support for More Models and Datasets}
Currently, TorchFL only supports major datasets that are oriented toward solving image recognition, classification, or computer vision tasks in general. The models we support are also state-of-the-art models which are used to solve image recognition and segmentation tasks. However, the design and interfaces that we discussed in Section \ref{subsection_datasets_models}, are backward compatible with a more diverse range of DL tasks including but not limited to natural language processing (NLP) or reinforcement learning (RL). As we plan to actively maintain \textit{TorchFL}, we are considering adding a diverse range of datasets and model implementations in the future. Again, it would only require the developers to make the new offerings backward compatible with the \textit{TorchFL} modules and it would automatically work with all the FL components.

\subsection{Broader Range of FL Components}
One of the primary motivations to design the federated modules as explained in Section \ref{subsection_federated} was to provide an easy-to-use interface to quickly prototype and validate the latest FL components. As the FL community is constantly innovating approaches toward sampling, agent incentivization, defense mechanisms, gradient/parameter encryption, and more, we plan to add those modules in addition to the samplers and aggregators that we currently support. In addition, we also aim to add more samplers and aggregator offerings which will serve as an example for the users who are willing to customize \textit{TorchFL} for their experiments.

\subsection{Benchmarking Against or Integrating with the Existing Toolkits}
In the previous section, we recognized various toolkits and the meaningful related works in the FL experiments and simulation space. One of our future goals is to benchmark \textit{TorchFL} against such toolkits in terms of performance, ease-of-use, models and datasets offerings, and more. As the DL community is making rapid progress, one of the major challenges is to keep up with state-of-the-art research, prototype the algorithm, and integrate it into the framework. As a result, one of our major goals is to get community feedback and support through our open-source offerings, and also focus on developer tooling and documentation which will hopefully make contributing to \textit{TorchFL} a lot easier. In addition to benchmarking, we believe that a lot of toolkits can also integrate and work with \textit{TorchFL} helping us to reduce the maintainable code. We are also open to focusing on such collaborations moving forward.




\bibliographystyle{plainnat} 
\bibliography{torchfl}

\newpage
\appendix
\section{Code Snippets}
This section will provide instructions and example code snippets for the users to get quickly started with \textit{TorchFL}. First, it will provide an example to use the \textit{datamodules} and \textit{models}, and later, it will demonstrate how to build on top of those to bootstrap an FL experiment.

\subsection{Using Datasets \& Models}
The following steps should be followed to bootstrap an experiment with EMNIST (MNIST) dataset and DenseNet121 model.

\begin{enumerate}
    \item Import the relevant modules as done in Figure \ref{fig:usage_1}.
\begin{figure}[ht!]
  \centering
  \includegraphics[keepaspectratio, width=0.7\columnwidth]{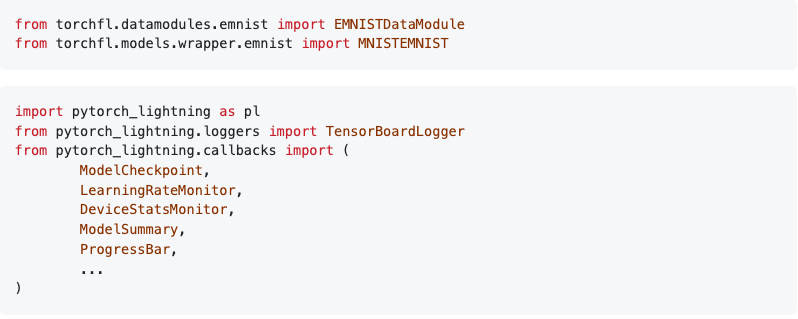}
  \caption{Import the relevant \textit{TorchFL} and PyTorch Lightning modules to get started. For more details, view the full list of PyTorch Lightning \href{https://pytorch-lightning.readthedocs.io/en/stable/extensions/callbacks.html\#callback}{callbacks} and \href{https://pytorch-lightning.readthedocs.io/en/latest/api_references.html\#loggers}{loggers}.}
  \label{fig:usage_1}
\end{figure}
    \item Set up the PyTorch Lightning Trainer API object as done in Figure \ref{fig:usage_2}.

\begin{figure}[ht!]
  \centering
  \includegraphics[keepaspectratio, width=0.7\columnwidth]{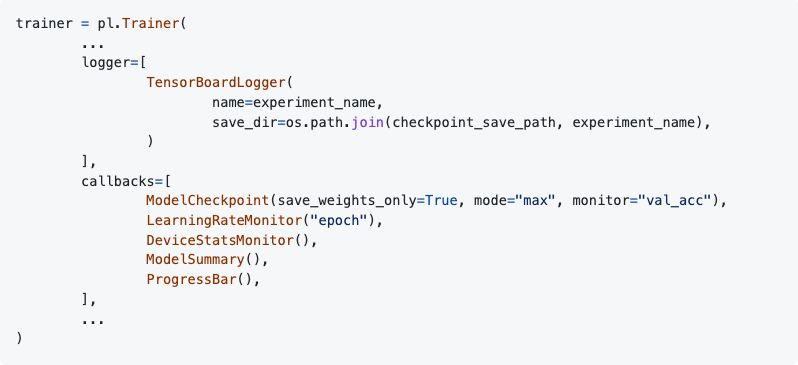}
  \caption{Set up the PyTorch Lightning Trainer object to initiate the training process. More details about the PyTorch Lightning \href{https://pytorch-lightning.readthedocs.io/en/latest/common/trainer.html}{Trainer API} can be found on their official website.}
  \label{fig:usage_2}
\end{figure}
    
    \item Prepare the dataset and model using the wrappers provided by the \textit{TorchFL}'s \textit{datamodules} and \textit{models} as shown in Figure \ref{fig:usage_3}.

\begin{figure}[ht!]
  \centering
  \includegraphics[keepaspectratio, width=0.7\columnwidth]{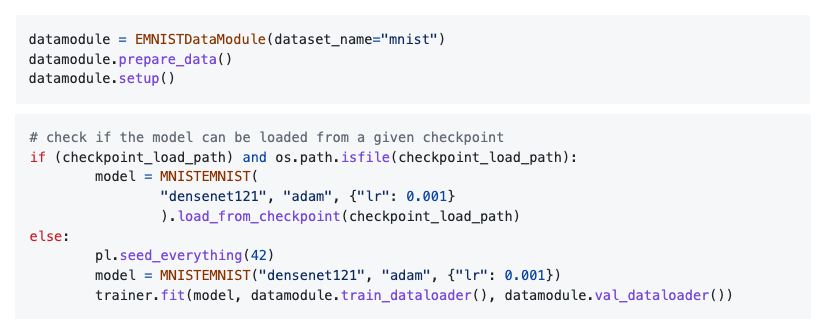}
  \caption{Prepare the dataset and model using \textit{TorchFL}'s \textit{datamodules} and \textit{models} wrappers.}
  \label{fig:usage_3}
\end{figure}

    \item The corresponding files for the experiment (checkpoints, metadata, etc.) will be stored at the \textit{default\_root\_dir} argument given to the PyTorch Lightning Trainer object in Step 2. For this experiment, we use the \href{https://www.tensorflow.org/tensorboard}{Tensorboard} logger. To view the logs (and related plots and metrics), go to the \textit{default\_root\_dir} path and find the Tensorboard log files. Upload the files to the Tensorboard Development portal following the instructions \href{https://tensorboard.dev/#get-started}{here}. Note that, \textit{TorchFL} is compatible with all the loggers supported by PyTorch Lightning. More information about the Lightning loggers can be found \href{https://tensorboard.dev/experiment/Q1tw19FySLSjLN6CW5DaUw/}{here}.

    \item More example scripts with various models and datasets can be found \href{https://github.com/vivekkhimani/torchfl/tree/master/examples/trainers}{here}.
\end{enumerate}

\subsection{Federated Learning}
Using the previously set up dataset and models, an FL experiment can be set up in the following manner.

\begin{enumerate}
    \item Use the dataset generated by \textit{TorchFL}'s \textit{datamodules} to create federated data shards with IID or non-IID distribution as shown in Figure \ref{fig:usage_4}.

\begin{figure}[ht!]
  \centering
  \includegraphics[keepaspectratio, width=0.7\columnwidth]{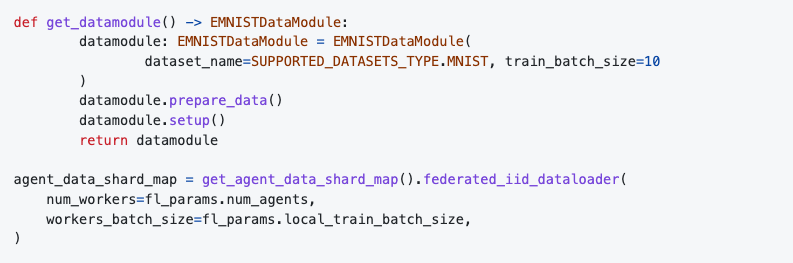}
  \caption{Create the data shards for the agents using the previously initialized dataset and model.}
  \label{fig:usage_4}
\end{figure}

\item Initialize a global model, and agents, and distribute the global model parameters to every agent as shown in Figure \ref{fig:usage_5}.

\begin{figure}[ht!]
  \centering
  \includegraphics[keepaspectratio, width=0.7\columnwidth]{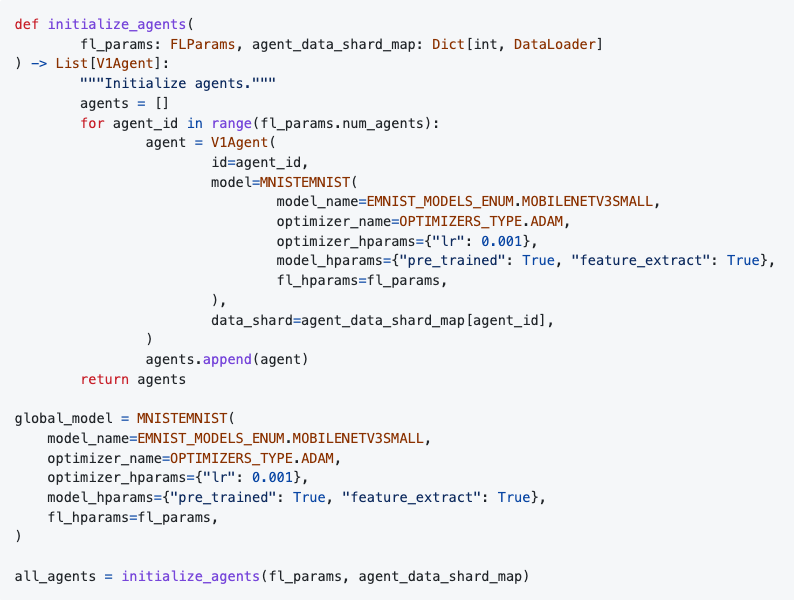}
  \caption{Initialize a global model, and  agents, and distribute the global model parameters to every agent.}
  \label{fig:usage_5}
\end{figure}

\item Initialize a \textit{TorchFL} \textit{FLParam} object with the desired FL hyperparameters and pass it on to the \textit{Entrypoint} object which will abstract the training as shown in Figure \ref{fig:usage_6}.

\begin{figure}[ht!]
  \centering
  \includegraphics[keepaspectratio, width=0.7\columnwidth]{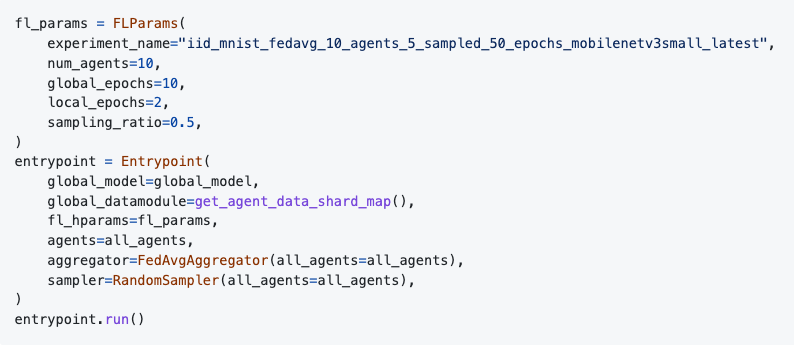}
  \caption{Initialize a \textit{TorchFL} \textit{FLParam} object with the desired FL hyperparameters and pass it on to the \textit{Entrypoint} object.}
  \label{fig:usage_6}
\end{figure}

\item More federated learning example scripts can be found \href{https://github.com/vivekkhimani/torchfl/tree/master/examples/federated}{here}.

\end{enumerate}


\end{document}